\documentclass[runningheads]{llncs}

\usepackage{accv}

\usepackage{accvabbrv}
\usepackage{graphicx}
\usepackage{booktabs}
\usepackage{multirow}
\usepackage{makecell}
\usepackage{pifont}
\usepackage{algorithm}
\usepackage[noend]{algorithmic}
\usepackage{wrapfig}

\usepackage[accsupp]{axessibility}

\newcommand{\argmax}{\mathop{\rm arg~max}\limits}

\usepackage{hyperref}

\usepackage{orcidlink}

\begin{document}

\title{Informative Viewpoint Selection for Episodic-Memory Embodied Question Answering using Omnidirectional Images}

\titlerunning{Informative Viewpoint Selection for EM-EQA using Omnidirectional Images}

\author{Kaname Kitamura\inst{1}\orcidlink{0009-0007-1068-5444} \and
Asako Kanezaki\inst{1,2,3}\orcidlink{0000-0003-3217-1405}}
\authorrunning{K. Kitamura and A. Kanezaki}
\institute{Department of Computer Science, Institute of Science Tokyo, Tokyo, Japan\\
\email{kitamura.k.c2ea@m.isct.ac.jp} \and
RIKEN, Japan \and
Tohoku University, Sendai, Japan\\
\email{asako.kanezaki.d4@tohoku.ac.jp}}

\maketitle

\begin{abstract}
Embodied Question Answering (EQA) requires agents to answer natural language questions about surrounding environments from visual observations. In this work, we focus on open-vocabulary episodic-memory EQA (EM-EQA), where an agent answers free-form questions using recorded observation histories. Omnidirectional images are promising for this task, as they provide wide field-of-view observations that can capture surrounding context without requiring explicit camera rotations. However, omnidirectional images introduce two challenges for EQA: (i) equirectangular projection causes severe geometric distortion that degrades vision-language model (VLM) recognition accuracy, and (ii) feeding equirectangular images directly into VLMs introduces excessive irrelevant background information, reducing answer accuracy and increasing the visual-token burden. To address these challenges, we propose a viewpoint selection method for EM-EQA using omnidirectional images. Our method converts equirectangular observations into perspective views via cubemap projection, estimates question-conditioned relevance with fine-tuned BLIP-2, and selects informative and diverse viewpoints through diversity-aware greedy selection. Experiments on the Habitat-Matterport 3D (HM3D) subset of OpenEQA show that our method achieves state-of-the-art model performance among the reported model results with equirectangular observations. Moreover, after removing rotation views, which reduces observation frames by 65.5\%, our method largely maintains its answer accuracy.
\end{abstract}

\section{INTRODUCTION}

\begin{figure}[t]
  \centering
  \includegraphics[width=\linewidth]{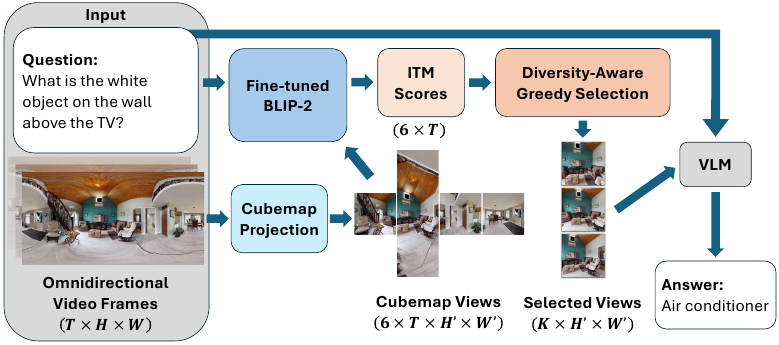}
  \caption{Overview of our viewpoint selection method for EM-EQA using omnidirectional images. Given an episode history of omnidirectional images and a natural language question, we convert images to perspective views via cubemap projection, compute relevance scores using fine-tuned BLIP-2, and select diverse informative viewpoints through our Diversity-Aware Greedy Selection algorithm for input to the VLM.}
  \label{fig:teaser}
\end{figure}

Embodied AI has made substantial progress toward agents that can navigate and interact with real-world environments, enabling complex tasks such as object manipulation~\cite{brohan2023rt1}, embodied reasoning~\cite{driess2023palme}, and task planning~\cite{ahn2022can}. Among these capabilities, Embodied Question Answering (EQA) has emerged as a fundamental benchmark for evaluating an agent's ability to understand its environment and communicate with humans through natural language~\cite{das2018embodied}.
EQA is commonly studied in two settings: episodic-memory EQA (EM-EQA), which uses recorded observation histories, and Active EQA (A-EQA), which requires agents to explore before answering.

The development of pretrained vision-language models (VLMs) has further shifted EQA toward open-vocabulary settings, where agents generate free-form answers rather than selecting from predefined answer candidates.
Omnidirectional images are a promising sensing modality for EQA because they provide wide field-of-view observations at each visited location, potentially reducing the need for explicit rotation actions and improving memory coverage.
However, directly applying omnidirectional images to EQA presents unique challenges:
\begin{itemize}
\item \textbf{Equirectangular Distortion:} Omnidirectional images are typically stored in equirectangular projection, which maps spherical coordinates to a 2D plane. This projection introduces severe geometric distortion, particularly near the poles where objects appear stretched and deformed~\cite{eder2020tangent, liu2025360}. Since mainstream VLMs are predominantly trained on standard perspective images~\cite{radford2021learning, zhang2024mmllms}, they struggle to correctly interpret equirectangular content, leading to degraded recognition accuracy.
\item \textbf{Information Overload:} An omnidirectional image covers a much wider field of view than a perspective image and therefore contains substantial question-irrelevant visual content. Feeding such observations directly into a VLM can distract the model from question-relevant evidence, reducing answer accuracy and increasing the visual-token burden~\cite{cheng2025efficienteqa}.
\end{itemize}

As summarized in \cref{fig:teaser}, we propose an informative viewpoint selection framework for EM-EQA using omnidirectional observations.
Our main contributions are summarized as follows:
\begin{itemize}
    \item We investigate open-vocabulary EM-EQA using omnidirectional images, a largely underexplored setting in prior EQA research. We highlight the potential of omnidirectional observations as a wide field-of-view sensing modality for improving spatial coverage in recorded observation histories.
    \item We propose an informative viewpoint selection framework for EM-EQA using omnidirectional images. Our method converts equirectangular observations into cubemap-based perspective views, estimates image-question relevance with BLIP-2 fine-tuned on filtered VQAv2 via supervised contrastive learning, and selects informative yet non-redundant viewpoints through Diversity-Aware Greedy Selection.
    \item We conduct experiments on the Habitat-Matterport 3D (HM3D) subset of the OpenEQA benchmark~\cite{majumdar2024openeqa} under full and rotation-reduced observation histories, showing that our method achieves state-of-the-art model performance among the reported model results with equirectangular observations.
    \item We demonstrate that, on the HM3D subset of OpenEQA, removing rotation views reduces the average number of observation frames by 65.5\%, from 170 to 59 per question, while our method largely maintains its answer accuracy.
\end{itemize}

\section{RELATED WORK}

\noindent \textbf{Embodied Question Answering.}
Embodied Question Answering (EQA) was first introduced by Das et al.~\cite{das2018embodied} as a task requiring agents to navigate 3D environments and answer questions based on visual observations. Unlike traditional Visual Question Answering (VQA)~\cite{antol2015vqa, goyal2017making} that operates on static images, EQA requires exploration and integration of observations across time and space.
Early EQA methods employed end-to-end reinforcement learning~\cite{das2018embodied, gordon2018iqa, yu2019multi}, training agents to simultaneously learn navigation policies and question answering. However, these approaches were limited to closed vocabularies and specific training environments, exhibiting poor generalization to novel questions or scenes.
The emergence of large-scale pretrained VLMs has enabled open-vocabulary EQA~\cite{majumdar2024openeqa, cheng2025efficienteqa}, where agents can generate free-form answers without task-specific training. OpenEQA~\cite{majumdar2024openeqa} established a comprehensive benchmark distinguishing between EM-EQA, where agents answer from recorded observations, and A-EQA, where agents actively explore to find answers.
Retrieval-augmented generation (RAG) has been adopted to handle long observation histories by retrieving question-relevant frames~\cite{cheng2025efficienteqa}, frame captions~\cite{ong2025reqa}, or entries from spatio-temporal and hierarchical memories~\cite{anwar2025remembr, xie2024embodiedrag}. Other methods use semantic maps~\cite{chen2023nlmap, huang2023vlmaps}, structured 3D memory~\cite{yang20253dmem}, dynamic scene-graph updates~\cite{ali2025graphpad}, or question-conditioned region relevance for efficient exploration~\cite{zhang2026fasteqa}. These methods mainly assume standard perspective observations or explicit 3D representations. In contrast, our work selects informative perspective views derived from omnidirectional observation histories and passes the selected images, rather than textual descriptions or memory entries, directly to a VLM for EM-EQA.

\noindent \textbf{Omnidirectional Image Processing.}
Omnidirectional images capture a holistic view of the surroundings but exhibit inherent geometric distortion when represented in equirectangular format~\cite{chou2020visual, eder2020tangent}. The distortion is most severe near the poles, where horizontal lines curve and objects appear stretched, making direct interpretation difficult for VLMs trained primarily on perspective images.
Several approaches have been developed to handle this distortion. Spherical CNNs~\cite{cohen2018spherical} perform convolutions directly on the sphere using spherical harmonics, tangent plane projections~\cite{eder2020tangent} approximate local regions with perspective patches, and other specialized architectures~\cite{li2023s2net, liu2025360} learn distortion-aware features for tasks such as depth estimation and semantic segmentation.
However, these methods often require additional training on omnidirectional images and cannot directly leverage strong pretrained VLMs. Projection-based approaches offer a practical alternative: gnomonic projection maps spherical regions onto tangent planes to produce perspective views~\cite{eder2020tangent}, and cubemap projection~\cite{chou2020visual} applies this principle to six cube faces, converting an omnidirectional image into six reduced-distortion perspective views compatible with standard models. We adopt cubemap projection to enable the use of pretrained VLMs while mitigating equirectangular distortion.
Omnidirectional images have been applied to VQA tasks, but only in limited, non-embodied settings. VQA-360~\cite{chou2020visual} and Pano-AVQA~\cite{yun2021pano} both target QA on individual omnidirectional images or videos, without considering embodied navigation. These differences make EM-EQA using omnidirectional images a distinct setting.

\noindent \textbf{Vision Language Models.}
Recent VLMs provide strong open-vocabulary visual understanding and answer-generation capabilities~\cite{openai2023gpt4, liu2023llava, bai2024qwen2vl}. In this work, we use BLIP-2~\cite{li2023blip2} for question-conditioned relevance scoring because its Image-Text Matching (ITM) head provides explicit image-text relevance scores. We use Qwen2-VL~\cite{bai2024qwen2vl} as the main answer model because it supports high-resolution visual inputs.

\section{PROPOSED METHOD}

\begin{figure}[t]
  \centering
  \begin{subfigure}[t]{0.49\linewidth}
    \centering
    \includegraphics[width=\linewidth]{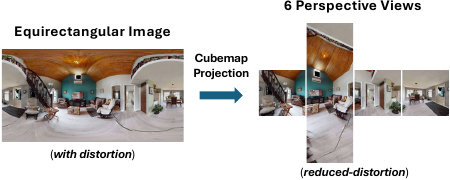}
    \caption{Cubemap projection}
    \label{fig:cubemap}
  \end{subfigure}
  \hfill
  \begin{subfigure}[t]{0.49\linewidth}
    \centering
    \includegraphics[width=\linewidth]{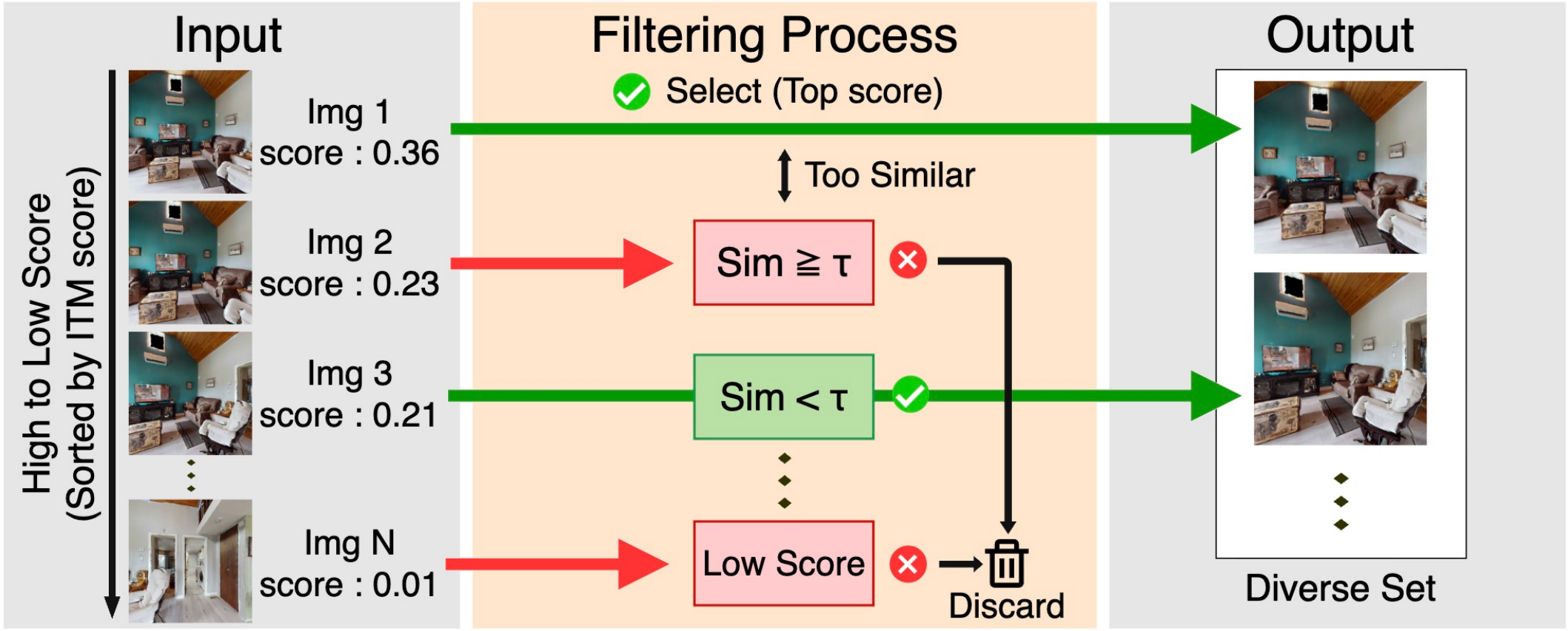}
    \caption{Diversity-Aware Greedy Selection}
    \label{fig:greedy}
  \end{subfigure}
  \caption{Cubemap projection converts an equirectangular image into six perspective views, and Diversity-Aware Greedy Selection selects informative viewpoints.}
  \label{fig:cubemap_greedy}

\end{figure}
\begin{figure}[t]
  \centering
  \includegraphics[width=\linewidth]{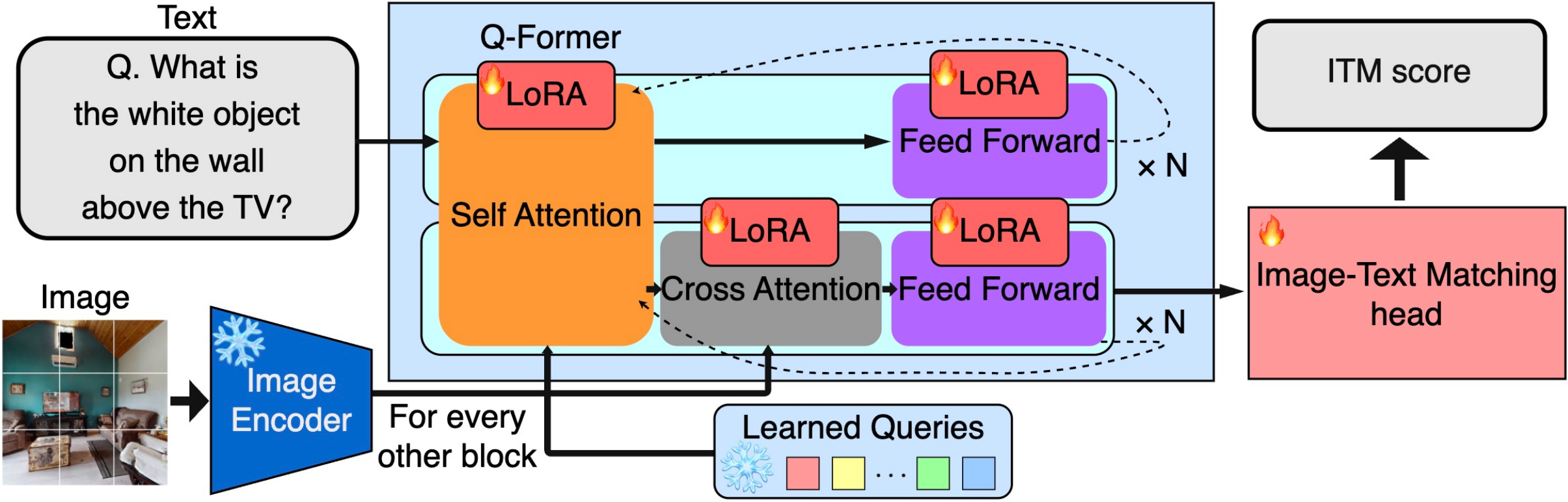}
  \caption{Architecture of fine-tuned BLIP-2 for relevance estimation.}
  \label{fig:blip2}
\end{figure}

\subsection{Problem Formulation}

Let $E = \{I_1, I_2, \ldots, I_T\}$ denote an episode history of $T$ equirectangular images of size $W \times H$ captured during agent navigation, and let $Q$ denote a natural language question. The objective of EM-EQA is to generate a natural language answer $A$ that correctly addresses $Q$ based on the recorded visual information in $E$.
Although VLMs can process multiple images, passing the entire recorded visual history is often undesirable because long histories may introduce irrelevant visual context into answer generation, exceed token limits, and require substantial computation. We therefore define the selected viewpoint subset as $\mathcal{S}^*$ and cast viewpoint selection as an optimization problem that seeks the visual evidence most useful for answering the question:
\begin{equation}
    \mathcal{S}^* = \argmax_{\mathcal{S} \subseteq \mathcal{V},\; |\mathcal{S}| = K} P(A^* \mid \mathcal{S}, Q),
\end{equation}
where $A^*$ is the correct answer, $\mathcal{V}$ denotes the candidate set of perspective viewpoints derived from $E$ through cubemap projection, $K$ is the number of selected viewpoints, $\mathcal{S}$ denotes a candidate subset of selected viewpoints, and $P(A^* \mid \mathcal{S}, Q)$ represents the ideal probability of producing the correct answer from the selected viewpoints and the question.

This objective motivates two design goals under a limited visual input budget: selected viewpoints should be relevant to the question and should avoid excessive redundancy so that they can cover complementary evidence. We implement these goals through question-conditioned relevance scoring and a diversity constraint.

\subsection{Cubemap Projection}

Equirectangular images exhibit severe geometric distortion that degrades VLM performance. We apply cubemap projection~\cite{chou2020visual} to decompose each omnidirectional image into six reduced-distortion perspective views corresponding to the faces of a cube: Front, Back, Left, Right, Top, and Bottom (\cref{fig:cubemap}).

We describe the backward sampling formulation used for resampling: for each pixel in an output cubemap face, we compute the corresponding location in the source equirectangular image. We first map pixel coordinates $(i, j)$ to relative coordinates $(a, b) \in [-1, 1]^2$, then compute the 3D viewing direction. For the Front face, the direction is $\mathbf{v} = (a, b, 1)$; the other five faces are defined by choosing the corresponding signed coordinate axis as the face normal and permuting $(a, b)$ accordingly.
After normalizing to unit vector $\hat{\mathbf{v}} = (\hat{x}, \hat{y}, \hat{z}) = \mathbf{v}/\|\mathbf{v}\|$, we convert to spherical coordinates:
\begin{equation}
    \theta = \mathrm{atan2}(\hat{x}, \hat{z}), \quad \phi = \arcsin(\hat{y}).
\end{equation}
Here, $\theta$ and $\phi$ denote the longitude and latitude angles on the viewing sphere, respectively.
Finally, we sample from the source equirectangular image at normalized horizontal and vertical coordinates $(u_{\mathrm{eq}}, v_{\mathrm{eq}}) \in [0, 1]^2$:
\begin{equation}
    u_{\mathrm{eq}} = \frac{\theta}{2\pi} + 0.5, \quad v_{\mathrm{eq}} = 0.5 - \frac{\phi}{\pi}.
\end{equation}
This transformation produces $6T$ candidate perspective viewpoints $\mathcal{V}$ from $T$ original omnidirectional images, each with reduced equirectangular distortion.

\subsection{Fine-tuned BLIP-2 for Relevance Estimation}

\begin{algorithm}[t]
\caption{Diversity-Aware Greedy Selection}
\label{alg:greedy}
\begin{algorithmic}[1]
\REQUIRE candidates $\mathcal{V}$, question $Q$, limit $K$, score threshold $\gamma$, similarity threshold $\tau$
\ENSURE selected viewpoints $\mathcal{S}$
\STATE $s_i \leftarrow f_{\mathrm{BLIP\text{-}2}}(I_i, Q)$ for all $I_i \in \mathcal{V}$
\STATE $\mathcal{C}_{\mathrm{high}} \leftarrow \{I_i \mid s_i \geq \gamma\}$; $\mathcal{C}_{\mathrm{low}} \leftarrow \{I_i \mid s_i < \gamma\}$
\IF{$\lvert\mathcal{C}_{\mathrm{high}}\rvert = 0$} \RETURN UniformSample ($\mathcal{V}$, $K$) \ENDIF
\STATE Sort $\mathcal{C}_{\mathrm{high}}$ by $s_i$ in descending order; $\mathcal{S} \leftarrow \emptyset$, $\mathcal{R} \leftarrow \emptyset$
\FORALL{$I_c \in \mathcal{C}_{\mathrm{high}}$}
    \IF{$\mathcal{S} = \emptyset$ \textbf{or} $\max_{I_s \in \mathcal{S}} \mathrm{sim}(\mathbf{e}_c, \mathbf{e}_s) < \tau$}
        \STATE $\mathcal{S} \leftarrow \mathcal{S} \cup \{I_c\}$
    \ELSE
        \STATE $\mathcal{R} \leftarrow \mathcal{R} \cup \{I_c\}$
    \ENDIF
    \IF{$\lvert\mathcal{S}\rvert \geq K$} \RETURN $\mathcal{S}$ \ENDIF
\ENDFOR
\IF{$\lvert\mathcal{S}\rvert < K$}
    \STATE Fill from $\mathcal{R}$, then $\mathcal{C}_{\mathrm{low}}$ with diversity constraint
\ENDIF
\RETURN $\mathcal{S}$
\end{algorithmic}
\end{algorithm}

We estimate image-question relevance using BLIP-2's ITM head~\cite{li2023blip2} (\cref{fig:blip2}). The pretrained BLIP-2 model is trained on image-caption pairs and is suboptimal for question-conditioned viewpoint selection. We fine-tune it using VQA data to improve relevance estimation.

For the training data, we use VQAv2~\cite{goyal2017making}, which contains 443K image-question pairs across 82K images. VQAv2 is designed with complementary image pairs where identical questions yield different answers, encouraging models to attend to image-specific details.
However, VQAv2 contains generic questions (e.g., ``What is in the picture?'') that can match many images and provide weak supervision for question-specific relevance estimation. We filter these questions using large language model (LLM)-based classification: questions classified as GENERIC are manually verified and removed, yielding 441K specific questions for training.

The ITM head performs binary classification on Q-Former output embeddings. We define the ITM score as the difference between match and mismatch logits: $s_{ij} = l^{(1)}_{ij} - l^{(0)}_{ij}$, where $l^{(1)}_{ij}$ and $l^{(0)}_{ij}$ are the match and mismatch logits for image $i$ and text $j$, respectively.
For a mini-batch of $B$ image-question pairs, let $t_i$ denote the question text paired with the $i$-th image. We define a text-matching mask matrix $\mathbf{M}$ with entries $M_{ij} = \delta(t_i, t_j)$, where $\delta(t_i, t_j) = 1$ if the two question texts are identical and 0 otherwise.
We use a supervised contrastive loss~\cite{khosla2020supervised} with the text-matching mask so that image-question pairs sharing the same question text are treated as positives within a mini-batch. The image-to-text loss is
\begin{equation}
    \mathcal{L}_{\mathrm{i2t}} = -\frac{1}{B} \sum_{i=1}^{B} \log \frac{\sum_{j} M_{ij} \exp(s_{ij})}{\sum_{j} \exp(s_{ij})}.
\end{equation}
The symmetric text-to-image loss $\mathcal{L}_{\mathrm{t2i}}$ is defined by swapping images and texts. The final loss $\mathcal{L}_{\mathrm{ITM}} = (\mathcal{L}_{\mathrm{i2t}} + \mathcal{L}_{\mathrm{t2i}})/2$ trains bidirectional image-text alignment.

We apply Low-Rank Adaptation (LoRA)~\cite{hu2021lora} to selected linear layers in the Q-Former, including attention projection and feed-forward layers, for efficient fine-tuning. LoRA adds low-rank matrices to pretrained weights $\mathbf{W}_0$:
\begin{equation}
    \mathbf{h} = \mathbf{W}_0 \mathbf{x} + \frac{\alpha}{r} \mathbf{W}_B \mathbf{W}_A \mathbf{x},
\end{equation}
where $\mathbf{x}$ is the input activation to the adapted linear layer, $\mathbf{h}$ is the resulting hidden representation, $\mathbf{W}_0 \in \mathbb{R}^{d \times k}$, $d$ and $k$ are the output and input dimensions of the layer, $\mathbf{W}_B \in \mathbb{R}^{d \times r}$, $\mathbf{W}_A \in \mathbb{R}^{r \times k}$, $r \ll \min(d, k)$, and $\alpha$ is the LoRA scaling factor. The image encoder and learned queries remain frozen, while the ITM head is fully fine-tuned. Training details are provided in \cref{sec:experiments}.

\subsection{Diversity-Aware Greedy Selection}

Naively selecting candidates in descending order of ITM score often produces visually similar viewpoints, wasting VLM capacity. We propose Diversity-Aware Greedy Selection to balance relevance and diversity, as shown in \cref{fig:greedy,alg:greedy}.

\noindent \textbf{Score-based Partitioning:} For each candidate $I_i$, the fine-tuned BLIP-2 computes the ITM score $s_i = f_{\mathrm{BLIP\text{-}2}}(I_i, Q)$, and we partition the candidates into two groups:

\begin{itemize}
    \item High-confidence set: $\mathcal{C}_{\mathrm{high}} = \{I_i \mid s_i \geq \gamma\}$
    \item Low-confidence set: $\mathcal{C}_{\mathrm{low}} = \{I_i \mid s_i < \gamma\}$
\end{itemize}
This partition prevents low-confidence candidates from being selected only because they are visually diverse. We first prioritize candidates that are sufficiently relevant to the question and use low-confidence candidates only as a fallback when the high-confidence set cannot provide enough diverse views.

\noindent \textbf{Greedy Selection with Diversity Constraint:} We iterate through $\mathcal{C}_{\mathrm{high}}$ in descending score order. A candidate $I_c$ is added to the selected set $\mathcal{S}$ only if its maximum similarity to existing selections is below the threshold $\tau$:
\begin{equation}
    \max_{I_s \in \mathcal{S}} \mathrm{sim}(\mathbf{e}_c, \mathbf{e}_s) < \tau,
\end{equation}
where $I_s$ denotes an already selected viewpoint in $\mathcal{S}$, $\mathrm{sim}(\cdot, \cdot)$ denotes cosine similarity, and $\mathbf{e}_c$ and $\mathbf{e}_s$ are the BLIP-2 image embeddings of $I_c$ and $I_s$, respectively. Each embedding $\mathbf{e} \in \mathbb{R}^{768}$ is computed as the mean of BLIP-2's 32 learned query embeddings for the image.

\noindent \textbf{Fallback Mechanisms:} To guarantee exactly $K$ outputs, if diversity constraints prevent reaching $K$, frames rejected from $\mathcal{C}_{\mathrm{high}}$, denoted by $\mathcal{R}$ in \cref{alg:greedy}, are added by score priority. If still insufficient, candidates from $\mathcal{C}_{\mathrm{low}}$ are added with diversity constraints. If $\lvert\mathcal{C}_{\mathrm{high}}\rvert = 0$, we fall back to uniform temporal sampling, as ITM-based selection is deemed unreliable.

\section{EXPERIMENTS}
\label{sec:experiments}

\subsection{Experimental Setup}

\noindent \textbf{Dataset:} OpenEQA~\cite{majumdar2024openeqa} provides a benchmark for open-vocabulary EQA in indoor environments. We use the EM-EQA setting on HM3D scenes~\cite{ramakrishnan2021hm3d}, where agents answer questions based on recorded observation histories.

We evaluate our method on all 557 HM3D questions within OpenEQA~\cite{majumdar2024openeqa}, covering seven question categories. We focus on the HM3D subset because the ScanNet~\cite{dai2017scannet} meshes in OpenEQA often lack regions uncaptured by the original cameras, producing large black voids in rendered equirectangular images.

For omnidirectional image experiments, we render equirectangular images at $2048 \times 1024$ resolution using the agent's position and orientation data provided by OpenEQA. For comparison, we also evaluate on standard perspective images at $1920 \times 1080$ resolution.

\noindent \textbf{Evaluation Metric:} To evaluate EQA accuracy, we employ the LLM-Match metric~\cite{majumdar2024openeqa}. Given a triplet of the question, reference answer, and model prediction, an LLM evaluator assigns a match score $\sigma_i \in \{1, 2, 3, 4, 5\}$. The final score is aggregated and normalized into a percentage:
\begin{equation}
  \mathit{LLM\text{-}Match\ Score} = \frac{1}{N} \sum_{i=1}^N \frac{\sigma_i - 1}{4} \times 100.
\end{equation}
Here, $N$ denotes the number of evaluated questions.

\noindent \textbf{Implementation Details:} For viewpoint selection, we use the BLIP-2 image-text matching model with the COCO variant from the LAVIS library. We apply LoRA with rank $r=16$ and scaling factor $\alpha=32$. Training uses the AdamW optimizer with a learning rate $10^{-4}$, a batch size of 16 for 10 epochs with 10\% warmup and cosine annealing. Experiments are conducted on NVIDIA H100 GPUs.

For answer generation, we use Qwen2-VL-72B-Instruct via Hugging Face Transformers. We evaluate across $K \in \{1, \dots, 10\}$ selected frames; experiments were not conducted for $K > 10$ because the token count would exceed the VLM input limit. We avoid image resizing because lower resolution may degrade recognition, especially for omnidirectional images where fine-grained objects often occupy small regions.
For LLM-Match evaluation, we use gpt-4-turbo-2024-04-09.

We compare our method with the following baselines:
\begin{itemize}
    \item \textbf{Multi-Frame VLMs}~\cite{majumdar2024openeqa}: Uniform temporal sampling of $K$ frames, the standard baseline for multi-frame VLM input.
    \item \textbf{Retrieval-Augmented Generation (RAG) with BLIP}: Following the RAG-based answering module of EfficientEQA~\cite{cheng2025efficienteqa}, this baseline selects the top-$K$ frames with the highest pretrained BLIP-2 ITM scores. For omnidirectional images, we evaluate two variants: (\textit{i}) Equirectangular, which applies the ITM score directly to the raw equirectangular images, and (\textit{ii}) Cubemap, which scores individual perspective views obtained via cubemap projection.
\end{itemize}

To assess applicability to standard perspective images, we also evaluate on standard images captured along the same agent trajectories. Here, our method omits cubemap projection and uses fine-tuned BLIP-2 with Diversity-Aware Greedy Selection for frame selection.

\begin{wraptable}[6]{r}{0.42\linewidth}
  \centering
\vspace{-35pt}
\caption{Average number of stored observation frames. Reduction rate is relative to \textit{Original}.}
  \label{tab:avg_obs_frames}
  \scriptsize
  \setlength{\tabcolsep}{2pt}
  \renewcommand{\arraystretch}{0.95}
  \begin{tabular}{@{}lcc@{}} \toprule
    Setting & Avg.\ frames & Reduction rate \\ \midrule
    Original & 170.48 & \textendash{} \\
    w/o rotation & 58.76 & 65.5\% \\ \bottomrule
  \end{tabular}
\end{wraptable}
We also test both omnidirectional and standard perspective observations after removing explicit rotation views. We compare two observation-history settings. The \textit{Original} setting uses the original OpenEQA trajectories with explicit rotational actions. The \textit{w/o rotation} setting uses the same trajectories but removes views from rotation actions, so the agent visits the same positions without explicit rotational views. This setting reduces the number of stored frames, as summarized in \cref{tab:avg_obs_frames}; this reduction results from removing rotation views from the observation history, not from our viewpoint selection. Unless otherwise stated, our method uses fixed thresholds $\tau=0.97$ and $\gamma=0.1$ across all conditions.

\subsection{Results and Analysis}
\begin{table}[t]
  \centering
  \caption{OpenEQA evaluation on EM-EQA in HM3D. Scores are maximum LLM-Match (\%) over $K \in \{1,\ldots,10\}$ for our experiments. Previously reported entries are quoted from the cited papers. Human baseline denotes the reported human performance, Blind LLMs denotes text-only models without visual input, CG denotes ConceptGraphs, and SVM denotes Sparse Voxel Map.}
  \label{tab:prior_work_comparison}
  \scriptsize
  \renewcommand{\arraystretch}{0.94}
  \setlength{\tabcolsep}{3.5pt}
  \begin{tabular*}{\linewidth}{@{\extracolsep{\fill}}llcc@{}} \toprule
    Method & Setting & Original & w/o rotation \\ \midrule
    Human baseline~\cite{majumdar2024openeqa} & perspective & 85.1 & \textendash \\
    GPT-4 (Blind LLMs)~\cite{majumdar2024openeqa} & \textendash & 35.5 & \textendash \\
    LLaMA-2 (Blind LLMs)~\cite{majumdar2024openeqa} & \textendash & 29.0 & \textendash \\
    GPT-4 w/ LLaVA-1.5~\cite{majumdar2024openeqa} & perspective & 40.0 & \textendash \\
    LLaMA-2 w/ LLaVA-1.5~\cite{majumdar2024openeqa} & perspective & 31.1 & \textendash \\
    GPT-4 w/ CG~\cite{majumdar2024openeqa} & perspective & 34.0 & \textendash \\
    LLaMA-2 w/ CG~\cite{majumdar2024openeqa} & perspective & 24.2 & \textendash \\
    GPT-4 w/ SVM~\cite{majumdar2024openeqa} & perspective & 35.0 & \textendash \\
    LLaMA-2 w/ SVM~\cite{majumdar2024openeqa} & perspective & 30.9 & \textendash \\
    GPT-4V (Multi-Frame VLMs)~\cite{majumdar2024openeqa} & perspective & 51.3 & \textendash \\
    AlanaVLM (Multi-Frame VLMs)~\cite{suglia2024alana} & perspective & 44.8 & \textendash \\
    Qwen2-VL (Multi-Frame VLMs) & perspective & 50.2 & 43.8 \\
    Qwen2-VL (RAG with BLIP) & perspective & 53.9 & 42.8 \\
    \textbf{Qwen2-VL (Ours)} & perspective & \textbf{61.1} & \textbf{45.5} \\ \midrule
    Qwen2-VL (Multi-Frame VLMs) & equirectangular & 54.5 & 52.6 \\
    Qwen2-VL (RAG with BLIP, equirectangular) & equirectangular & 50.4 & 51.3 \\
    Qwen2-VL (RAG with BLIP, cubemap) & equirectangular & 44.3 & 42.7 \\
    \textbf{Qwen2-VL (Ours)} & equirectangular & \textbf{57.9} & \textbf{55.9} \\ \bottomrule
  \end{tabular*}
\end{table}
\begin{table}[t]
  \centering
  \caption{Full LLM-Match scores (\%) over the number of selected frames $K$. Multi denotes Multi-Frame VLMs, RAG denotes RAG with BLIP, and bold indicates the highest score over $K$ for each method and observation setting.}
  \label{tab:detailed_main_results}
  \scriptsize
  \renewcommand{\arraystretch}{0.9}
  \setlength{\tabcolsep}{1.8pt}
  \resizebox{\linewidth}{!}{
  \begin{tabular}{@{}c*{14}{c}@{}} \toprule
    & \multicolumn{8}{c}{Equirectangular} & \multicolumn{6}{c}{Standard perspective} \\
    \cmidrule(lr){2-9}\cmidrule(l){10-15}
    & \multicolumn{4}{c}{Original} & \multicolumn{4}{c}{w/o rotation} & \multicolumn{3}{c}{Original} & \multicolumn{3}{c}{w/o rotation} \\
    \cmidrule(lr){2-5}\cmidrule(lr){6-9}\cmidrule(lr){10-12}\cmidrule(l){13-15}
    $K$ & Multi & \makecell{RAG} & \makecell{RAG\\cubemap} & Ours & Multi & \makecell{RAG} & \makecell{RAG\\cubemap} & Ours & Multi & RAG & Ours & Multi & RAG & Ours \\ \midrule
    1 & 35.0 & 49.7 & 41.7 & 50.5 & 34.7 & 49.4 & 37.4 & 51.3 & 30.1 & 44.9 & 52.7 & 30.6 & 37.5 & 41.8 \\
    2 & 45.6 & 49.2 & 43.7 & 53.4 & 45.4 & 49.1 & 40.2 & 54.4 & 35.8 & 47.4 & 54.9 & 38.2 & 37.1 & 45.0 \\
    3 & 51.4 & 48.2 & 43.8 & 57.0 & 49.6 & 51.1 & 40.4 & 54.9 & 44.5 & 49.2 & 57.2 & 42.0 & 38.0 & 44.9 \\
    4 & 50.8 & 48.7 & 42.1 & 55.7 & 50.9 & 50.7 & 40.8 & 55.3 & 43.9 & 50.1 & 59.4 & 41.1 & 39.1 & 44.7 \\
    5 & \textbf{54.5} & 49.9 & 41.7 & 57.5 & 50.9 & 50.4 & 42.6 & 55.4 & 47.1 & 50.9 & \textbf{61.1} & 42.9 & 40.1 & \textbf{45.5} \\
    6 & 53.5 & \textbf{50.4} & 42.6 & 57.5 & 51.7 & \textbf{51.3} & \textbf{42.7} & 55.2 & 46.0 & 51.1 & 60.2 & 43.4 & 40.6 & 45.0 \\
    7 & 52.8 & 49.3 & 42.8 & 57.5 & 51.0 & 50.0 & 41.4 & \textbf{55.9} & 48.7 & 52.2 & 60.6 & \textbf{43.8} & 41.6 & 44.6 \\
    8 & 54.4 & 49.4 & \textbf{44.3} & \textbf{57.9} & \textbf{52.6} & 50.0 & 42.6 & 55.4 & 46.8 & 52.6 & 60.5 & 40.9 & 41.8 & 44.4 \\
    9 & 52.9 & 48.7 & 43.8 & 57.0 & 51.3 & 50.4 & 42.3 & 55.1 & 49.8 & 53.5 & 59.7 & 43.3 & 41.7 & 43.6 \\
    10 & 53.5 & 48.8 & 43.8 & 55.9 & 51.4 & 51.0 & 42.4 & 55.5 & \textbf{50.2} & \textbf{53.9} & 59.8 & 42.6 & \textbf{42.8} & 43.9 \\ \bottomrule
  \end{tabular}
  }
\end{table}
\begin{table}[t]
  \centering
  \scriptsize
  \begin{minipage}[t]{0.61\linewidth}
    \centering
    \caption{Category-wise LLM-Match (\%) at $K=5$ on equirectangular observations in the \textit{Original} setting. OR, AR, OSR, OL, SU, FR, and WK denote Object Recognition, Attribute Recognition, Object State Recognition, Object Localization, Spatial Understanding, Functional Reasoning, and World Knowledge, respectively.}
    \label{tab:category_results}
    \setlength{\tabcolsep}{2pt}
    \begin{tabular}{@{}lcccccccc@{}} \toprule
      Method & OR & AR & OSR & OL & SU & FR & WK & All \\ \midrule
      Multi & 40.3 & 64.3 & \textbf{75.8} & \textbf{43.7} & 39.1 & 56.0 & \textbf{61.1} & 54.5 \\
      RAG   & 42.3 & 59.7 & 71.4 & 36.4 & 37.7 & 52.1 & 49.0 & 49.9 \\
      Ours  & \textbf{51.7} & \textbf{67.9} & \textbf{75.8} & 40.5 & \textbf{55.8} & \textbf{57.0} & 55.9 & \textbf{57.5} \\ \bottomrule
    \end{tabular}
  \end{minipage}\hfill
  \begin{minipage}[t]{0.36\linewidth}
    \centering
    \caption{LLM-Match (\%) with different answer models at $K=5$ on equirectangular observations in the \textit{Original} setting. Each answer model receives the same selected views for each method.}
    \label{tab:answer_backbones}
    \setlength{\tabcolsep}{2pt}
    \begin{tabular}{@{}lccc@{}} \toprule
      Answer model & Multi & RAG & Ours \\ \midrule
      Qwen2-VL-72B  & 54.5 & 49.9 & \textbf{57.5} \\
      GPT-5.6 Sol   & 56.9 & 52.9 & \textbf{66.6} \\
      InternVL3-78B & 51.9 & 49.7 & \textbf{58.0} \\
      Llama 4 Scout & 54.8 & 52.5 & \textbf{57.9} \\ \bottomrule
    \end{tabular}
  \end{minipage}
\end{table}

\Cref{tab:prior_work_comparison} summarizes our best results and OpenEQA~\cite{majumdar2024openeqa} scores on HM3D.
We report the maximum score over $K$ as a summary and provide the full sweep in \cref{tab:detailed_main_results}; in practice, $K$ can be set by the input budget of the answer model.
Our method achieves the highest score among the evaluated selection strategies across both observation types and both rotation settings, achieving state-of-the-art model performance among the reported model results.
Standard perspective observations obtain the highest overall score in the \textit{Original} setting, but degrade much more after removing views from rotation actions~(\cref{tab:avg_obs_frames}): our method drops by 15.6 points for standard perspective observations (61.1 to 45.5), whereas the drop is only 2.0 points for equirectangular observations (57.9 to 55.9).
This suggests that standard perspective observations rely heavily on views captured during rotation actions, while omnidirectional observations are less sensitive to the removal of such views in EM-EQA.

The results also reveal the limitations of pretrained BLIP-2 for retrieval over omnidirectional observations. For equirectangular observations, RAG with BLIP (50.4\% and 51.3\%) underperforms Multi-Frame VLMs in both settings, suggesting that distortion and wide visual coverage hinder relevance estimation. The cubemap-based variant performs even worse (44.3\% and 42.7\%), likely because the six-fold larger candidate pool makes retrieval more sensitive to irrelevant or redundant views without task-specific fine-tuning.

Across all observation settings and values of $K$ in \cref{tab:detailed_main_results}, our method consistently outperforms the compared methods under the same input-view budget and experimental setting.
Notably, our method reaches higher scores than the best baseline with fewer selected frames in all observation settings, indicating that the selected views contain more informative visual evidence for answering the question.
Interestingly, the best performance is not always obtained at the largest $K$.
In many settings, the scores plateau around $K=5$–$8$, and further increasing $K$ provides limited additional benefit.
This trend suggests that adding more frames can introduce redundant or less relevant observations while also increasing computational cost.

\Cref{tab:category_results} shows that our method achieves the best or tied-best score in five of the seven categories, with the largest gain in Spatial Understanding (55.8\% vs.\ 39.1\% for Multi-Frame VLMs and 37.7\% for RAG with BLIP).

\Cref{tab:answer_backbones} shows that, with GPT-5.6 Sol~\cite{openai2026gpt56}, InternVL3-78B~\cite{zhu2025internvl3}, or Llama 4 Scout~\cite{meta2025llama4} as the answer model, our method also outperforms Multi-Frame VLMs and RAG with BLIP. Implementation details of the selector and answer-model comparisons are given in the supplementary material.

\subsection{Ablation Studies}
\begin{table}[t]
  \centering
  \captionsetup{skip=0pt}
  \caption{Ablation studies at $K=5$ on equirectangular observations in the \textit{Original} setting. Scores are LLM-Match (\%). FT denotes Fine-tuning, and DAGS denotes Diversity-Aware Greedy Selection. R-EQA-style ranks views by caption similarity without BLIP-2, and MMR uses $\lambda=0.5$. The similarity sweep fixes $\gamma=0.1$, and the score sweep fixes $\tau=0.97$.}
  \captionsetup{skip=3pt}
  \vspace{-3mm}
  \scriptsize
  \renewcommand{\arraystretch}{0.92}
  \begin{subtable}[t]{0.37\linewidth}
    \centering
    \caption{Components and selectors}
    \label{tab:ablation_component}
    \setlength{\tabcolsep}{2.3pt}
    {\renewcommand{\arraystretch}{1.10}
    \begin{tabular}{@{}cccc@{}} \toprule
      Cubemap & FT & Selector & Score $\uparrow$ \\ \midrule
      \ding{55} & \ding{55} & Top-$K$ & 49.9 \\
      \ding{55} & \ding{51} & Top-$K$ & 51.4 \\
      \ding{55} & \ding{51} & DAGS & 53.7 \\
      \ding{51} & \ding{55} & Top-$K$ & 41.7 \\
      \ding{51} & \ding{51} & Top-$K$ & 55.2 \\
      \ding{51} & \textendash & R-EQA-style & 50.3 \\
      \ding{51} & \ding{51} & MMR & 55.8 \\
      \ding{51} & \ding{51} & DAGS & \textbf{57.5} \\ \bottomrule
    \end{tabular}
    }
  \end{subtable}
  \hfill
  \begin{subtable}[t]{0.60\linewidth}
    \centering
    \caption{Thresholds}
    \label{tab:ablation_tau}
    \label{tab:ablation_gamma}
    \setlength{\tabcolsep}{1.1pt}
    \textbf{Similarity threshold $\tau$}\\
    \begin{tabular*}{\linewidth}{@{\extracolsep{\fill}}l*{11}{c}@{}} \toprule
      $\tau$ & 0.90 & 0.91 & 0.92 & 0.93 & 0.94 & 0.95 & 0.96 & 0.97 & 0.98 & 0.99 & 1.00 \\ \midrule
      Score & 55.8 & 56.0 & 56.4 & 55.3 & 56.3 & 56.8 & 57.3 & \textbf{57.5} & 55.7 & 56.9 & 56.4 \\ \bottomrule
    \end{tabular*}\par\smallskip
    \textbf{Score threshold $\gamma$}\\
    \begin{tabular*}{0.72\linewidth}{@{\extracolsep{\fill}}l*{6}{c}@{}} \toprule
      $\gamma$ & 0.0 & 0.1 & 0.2 & 0.3 & 0.4 & 0.5 \\ \midrule
      Score & 56.1 & 57.5 & 57.7 & 57.1 & 57.5 & \textbf{57.9} \\ \bottomrule
    \end{tabular*}
  \end{subtable}
\end{table}

\noindent \textbf{Component Analysis:}
\Cref{tab:ablation_component} isolates the contribution of each component at $K=5$.
The baseline, corresponding to RAG with BLIP on equirectangular images, achieves 49.9\% with pretrained BLIP-2.
Cubemap projection alone decreases the score to 41.7\%, consistent with \cref{tab:prior_work_comparison}, whereas fine-tuning improves equirectangular scoring to 51.4\% and cubemap-based scoring to 55.2\%.
This shows that task-specific fine-tuning is crucial for identifying question-relevant views, especially among cubemap-projected candidates.
Adding Diversity-Aware Greedy Selection further improves the score to 57.5\%.
These results indicate complementary roles: fine-tuning improves relevance estimation, cubemap projection provides localized perspective views with reduced equirectangular distortion, and diversity-aware selection reduces redundancy, leading to a more informative set of selected views.

\noindent \textbf{Comparison with Other Selectors:}
As shown in \cref{tab:ablation_component}, Diversity-Aware Greedy Selection (57.5\%) outperforms maximal marginal relevance (MMR)~\cite{carbonell1998mmr} with the same fine-tuned BLIP-2 (at most 55.8\% over relevance weights $\lambda \in \{0.1,\ldots,0.9\}$) and an R-EQA-style selector~\cite{ong2025reqa} (50.3\%) under the same cubemap candidates, answer model, and $K$. Unlike MMR, which optimizes a weighted relevance--redundancy objective, our method preserves the relevance ranking and uses visual similarity as a constraint. Unlike the R-EQA-style selector, which ranks views by Sentence-BERT~\cite{reimers2019sentencebert} similarity between the question and view captions generated by Qwen2.5-VL-7B-Instruct~\cite{bai2025qwen25vl}, it scores views directly from images.

\noindent \textbf{Similarity Threshold $\tau$:}
\Cref{tab:ablation_tau} reports the similarity threshold sweep.
In this sweep, $\tau=0.97$ gives the highest score at $K=5$.

\noindent \textbf{Score Threshold $\gamma$:}
\Cref{tab:ablation_gamma} reports the score threshold sweep.
All tested settings with $\gamma \in [0.1, 0.5]$ improve over $\gamma=0.0$, indicating that score-based partitioning consistently benefits viewpoint selection.
We adopt $\gamma=0.1$ as the smallest such threshold to minimize candidate exclusion.

\subsection{Qualitative Analysis}

\begin{figure}[t]
  \centering
  \begin{subfigure}[t]{0.49\linewidth}
    \centering
    \includegraphics[width=\linewidth]{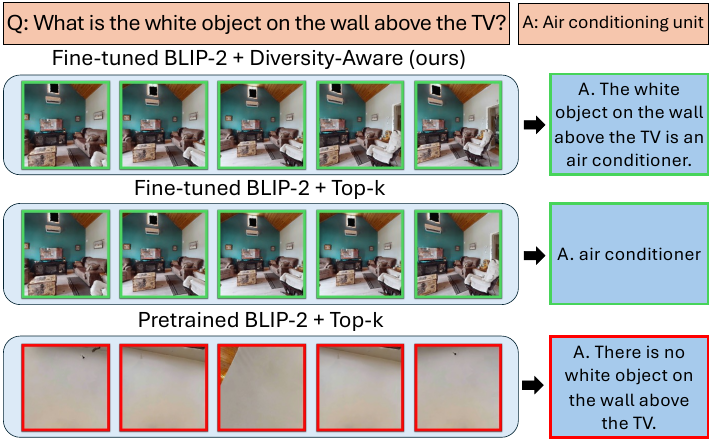}
    \caption{Target object is the air conditioner.}
    \label{fig:qual_ac}
  \end{subfigure}
  \begin{subfigure}[t]{0.49\linewidth}
    \centering
    \includegraphics[width=\linewidth]{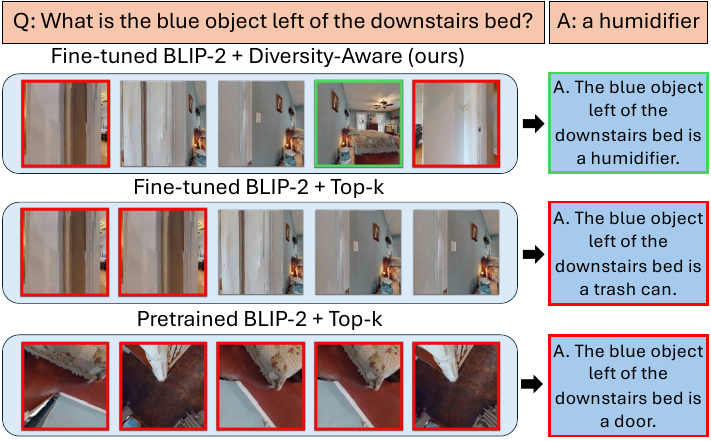}
    \caption{Target object is the humidifier.}
    \label{fig:qual_hum}
  \end{subfigure}
  \caption{Qualitative comparison of frame selection.}
  \label{fig:qualitative}
\end{figure}

\Cref{fig:qualitative} qualitatively evaluates the differences in viewpoint selection strategies on cubemap-projected perspective views.

In \cref{fig:qual_ac}, pretrained BLIP-2 assigns high ITM scores to irrelevant ceiling views, likely triggered by surface-level keyword matching with ``white'' and ``wall.'' In contrast, our fine-tuned model identifies frames in which the TV and target air conditioner are visible, enabling selection of the decisive viewpoint. This demonstrates that fine-tuning improves question-image relevance estimation.

In \cref{fig:qual_hum}, pretrained BLIP-2 similarly fails by selecting irrelevant frames.
While fine-tuning identifies frames containing the target object, simple Top-$K$ selection repeatedly picks similar partial views, causing the VLM to miss the spatial evidence needed for the correct answer. In contrast, Diversity-Aware Greedy Selection filters near-duplicate views and selects a viewpoint that captures the spatial relationship between the bed and the target object. With this selected evidence, the VLM generates the correct answer, suggesting that diversity-constrained selection provides more informative viewpoints for answer generation.

\subsection{Inference Time Analysis}
\label{sec:time}

\begin{table}[t]
  \centering
  \caption{Inference time components at $K=5$. Times are seconds per question. Selection denotes viewpoint selection time, Answer denotes answer generation time, and Total is their sum. Multi denotes Multi-Frame VLMs, RAG denotes RAG with BLIP, and RAG (cube) denotes cubemap-based RAG with BLIP.}
  \label{tab:time}
  \scriptsize
  \setlength{\tabcolsep}{1.8pt}
  \renewcommand{\arraystretch}{0.9}
  \resizebox{\linewidth}{!}{
  \begin{tabular}{@{}l*{14}{c}@{}} \toprule
    & \multicolumn{8}{c}{Equirectangular} & \multicolumn{6}{c}{Standard perspective} \\
    \cmidrule(lr){2-9}\cmidrule(l){10-15}
    & \multicolumn{4}{c}{Original} & \multicolumn{4}{c}{w/o rotation} & \multicolumn{3}{c}{Original} & \multicolumn{3}{c}{w/o rotation} \\
    \cmidrule(lr){2-5}\cmidrule(lr){6-9}\cmidrule(lr){10-12}\cmidrule(l){13-15}
    Stage & Multi & RAG & \makecell{RAG\\(cube)} & Ours & Multi & RAG & \makecell{RAG\\(cube)} & Ours & Multi & RAG & Ours & Multi & RAG & Ours \\ \midrule
    Selection & 0.00 & 9.54 & 16.56 & 25.23 & 0.00 & 3.47 & 5.59 & 9.40 & 0.00 & 8.64 & 11.50 & 0.00 & 3.14 & 4.11 \\
    Answer & 8.39 & 8.21 & 1.89 & 2.53 & 8.32 & 8.29 & 1.89 & 2.70 & 8.34 & 8.48 & 8.47 & 8.47 & 8.52 & 8.46 \\
    Total & 8.39 & 17.75 & 18.45 & 27.76 & 8.32 & 11.76 & 7.48 & 12.10 & 8.34 & 17.12 & 19.97 & 8.47 & 11.66 & 12.57 \\ \bottomrule
  \end{tabular}
  }
\end{table}

\Cref{tab:time} reports inference time at $K=5$. Our method incurs additional computation compared with Multi-Frame VLMs and RAG-based baselines because the viewpoint selection stage requires scoring candidate views and extracting image features for diversity-aware selection. For equirectangular observations, this cost is further increased by cubemap projection, which expands each omnidirectional observation into six perspective candidates before scoring. However, removing explicit rotation views reduces the equirectangular inference time from 27.76 to 12.10 seconds, mainly because the average number of observation frames decreases from 170 to 59 per question~(\cref{tab:avg_obs_frames}). Together with the accuracy results in \cref{tab:prior_work_comparison}, this indicates a favorable trade-off: the \textit{w/o rotation} setting reduces stored frames and inference cost while largely preserving answer accuracy and avoiding explicit rotational scanning.

This trade-off is meaningful because the two costs are reduced in different ways. The selection time reflects our current, unoptimized implementation, which repeatedly projects, scores, and encodes candidate views, and could be reduced by more efficient candidate processing or by reusing visual features accumulated during observation. In contrast, explicit rotation views require the camera or robot body to physically rotate during data collection, which software optimization alone cannot remove and which may add time, energy, and control burden. Thus, the benefit of the \textit{w/o rotation} setting goes beyond faster inference: it reduces dependence on repeated rotational scanning.

\section{LIMITATIONS}
\label{sec:limitations}

Our approach relies on a small set of independently selected recorded viewpoints, which may be insufficient for global scene understanding, temporal trajectory reasoning, multi-view integration, absence verification, counting, and multi-instance disambiguation. Moreover, ITM-based relevance and visual diversity do not directly measure whether the selected views provide complementary answer-bearing evidence. Structured memory mechanisms could address these cases by maintaining object identities, temporal context, and scene-level evidence.

Although cubemap projection reduces equirectangular distortion, objects near face boundaries can be split or truncated, and the six fixed cube faces may not match the optimal continuous viewing directions for each question. More adaptive projection strategies are therefore needed.

Our method improves answer accuracy but introduces additional computation due to cubemap-based candidate expansion, BLIP-2 scoring, and image-feature extraction for selection. This overhead could be reduced by a more efficient implementation (\cref{sec:time}).

Finally, our evaluation is limited to the HM3D subset of OpenEQA, and the full selection framework is evaluated only with BLIP-2 as the relevance model. Evaluating the full framework with other relevance models, in additional environments, and on real robots with omnidirectional cameras is future work.

\section{CONCLUSION}

We investigated EM-EQA using omnidirectional images.
To address equirectangular distortion and information overload, we proposed a viewpoint selection method that combines cubemap projection, question-conditioned relevance estimation, and diversity-aware selection.
On the HM3D subset of OpenEQA, our method achieves state-of-the-art model performance among the reported model results with equirectangular observations, and after removing rotation views, which reduces observation frames by 65.5\%, it largely maintains its answer accuracy.
These results suggest that our method can extract informative views for the VLM from omnidirectional observations, and that omnidirectional observation histories can provide useful scene coverage for EQA.

\clearpage

\bibliographystyle{splncs04}
\bibliography{main}

\clearpage
\setcounter{section}{0}
\setcounter{table}{0}
\setcounter{figure}{0}
\setcounter{equation}{0}
\renewcommand{\thesection}{S\arabic{section}}
\renewcommand{\thetable}{S\arabic{table}}
\renewcommand{\thefigure}{S\arabic{figure}}
\renewcommand{\theequation}{S\arabic{equation}}
\renewcommand{\theHsection}{S\arabic{section}}
\renewcommand{\theHtable}{S\arabic{table}}
\renewcommand{\theHfigure}{S\arabic{figure}}
\renewcommand{\theHequation}{S\arabic{equation}}
\setlength{\tabcolsep}{6pt}
\raggedbottom
\makeatletter
\setlength{\@fptop}{0pt}
\makeatother

\begin{center}
{\Large\bfseries Informative Viewpoint Selection for Episodic-Memory Embodied Question Answering using Omnidirectional Images\par}
\vskip 1em
{\large Supplementary Material\par}
\end{center}
\vskip 1em
\suppressfloats[t]

This supplementary material provides additional analyses and implementation details that complement the main paper. Unless otherwise stated, all results use equirectangular observations in the \textit{Original} setting of the Habitat-Matterport 3D (HM3D) subset of OpenEQA (557 questions), Qwen2-VL-72B-Instruct as the answer model, $K=5$, score threshold $\gamma=0.1$, and similarity threshold $\tau=0.97$. Scores are LLM-Match (\%) computed with gpt-4-turbo-2024-04-09, as in the main paper. Notation and abbreviations follow the main paper, and tables and algorithms without the prefix ``S'' refer to the main paper.

\section{Details of Additional Comparisons}

\noindent\textbf{MMR.}
Maximal marginal relevance (MMR)~\cite{carbonell1998mmr} uses the same cubemap candidates, fine-tuned BLIP-2 relevance, and BLIP-2 image embeddings as our method. Given the selected set $\mathcal{S}$, it iteratively selects
\begin{equation}
  I^{*} = \operatorname*{arg\,max}_{I_c \in \mathcal{V} \setminus \mathcal{S}} \Bigl[\lambda\, r_c - (1-\lambda) \max_{I_s \in \mathcal{S}} \mathrm{sim}(\mathbf{e}_c, \mathbf{e}_s)\Bigr],
\end{equation}
where $\mathcal{V}$ is the set of candidate views, $I_c$ and $I_s$ are a candidate view and a selected view, $\mathbf{e}_c$ and $\mathbf{e}_s$ are their BLIP-2 image embeddings, $r_c$ is the match probability of the image-text matching (ITM) head for $I_c$, $\mathrm{sim}(\cdot,\cdot)$ is cosine similarity, and $\lambda$ is the relevance weight. \Cref{tab:mmr} reports the results for $\lambda \in \{0.1, \ldots, 0.9\}$. The highest score is 55.8\% at $\lambda=0.5$.

\begin{table}[t]
  \centering
  \caption{MMR with different relevance weights $\lambda$.}
  \label{tab:mmr}
  \begin{tabular}{@{}l*{9}{c}@{}} \toprule
    $\lambda$ & 0.1 & 0.2 & 0.3 & 0.4 & 0.5 & 0.6 & 0.7 & 0.8 & 0.9 \\ \midrule
    LLM-Match & 50.6 & 51.7 & 54.4 & 55.3 & \textbf{55.8} & 54.5 & 53.8 & 54.0 & 54.2 \\ \bottomrule
  \end{tabular}
\end{table}

\noindent\textbf{R-EQA-style selector.}
This baseline follows the publicly described retrieval policy of R-EQA~\cite{ong2025reqa} and is not a full reproduction. Each $512 \times 512$ cubemap candidate is captioned with Qwen2.5-VL-7B-Instruct~\cite{bai2025qwen25vl}. Captions and the question are embedded with Sentence-BERT~\cite{reimers2019sentencebert} (\texttt{all-MiniLM-L6-v2}), and the top-5 views by cosine similarity are passed to the same answer model.

\noindent\textbf{Answer models.}
For \cref{tab:answer_backbones}, each answer model receives the same precomputed selections (the same five images in the same order) for each method, with the same prompt and a maximum of 128 generated tokens. GPT-5.6 Sol~\cite{openai2026gpt56} (\texttt{gpt-5.6-sol}) is accessed through the Responses API with temperature 0, reasoning effort \texttt{none}, and image detail \texttt{auto} at 512 pixels. InternVL3-78B~\cite{zhu2025internvl3} and Llama 4 Scout 17B-16E-Instruct~\cite{meta2025llama4} are run with temperature 0 and seed 1234.

\section{Qualitative Examples}

\begin{figure}[t]
  \centering
  \begin{subfigure}{\linewidth}
    \centering
    \includegraphics[width=\linewidth]{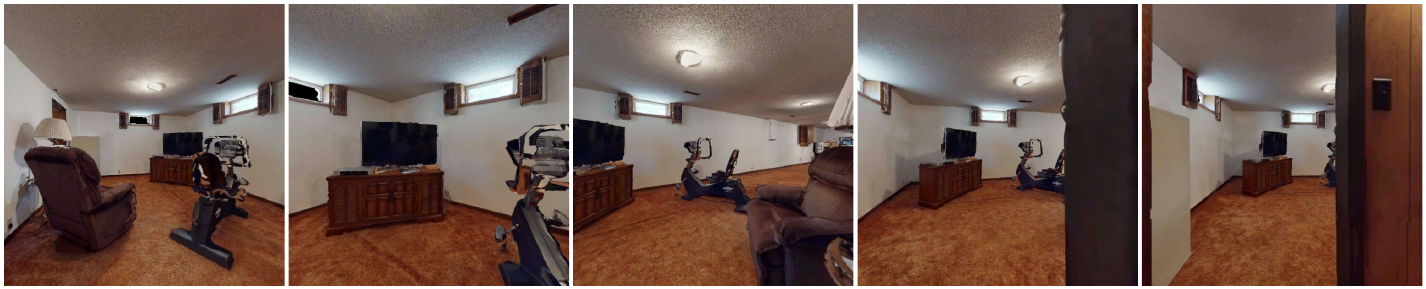}
    \caption{Implicit-target question ``I go to the TV room, which object can I use to exercise?''}
    \label{fig:implicit}
  \end{subfigure}\\[4pt]
  \begin{subfigure}{\linewidth}
    \centering
    \includegraphics[width=\linewidth]{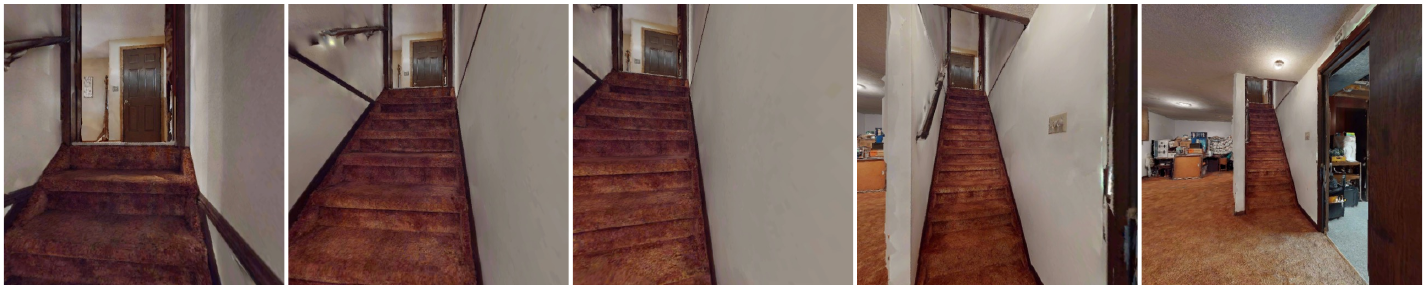}
    \caption{Route-following question ``If you were to go down the stairs, then straight into the tv room, turn left in the door and go straight until you enter the last room, which room would it be?''}
    \label{fig:route}
  \end{subfigure}
  \caption{Views selected by our method for two qualitative examples.}
  \label{fig:supp_qualitative}
\end{figure}

\noindent\textbf{Implicit target inference.}
\Cref{fig:implicit} shows an example in which the question does not name the target object. For the question ``I go to the TV room, which object can I use to exercise?'', ITM-based selection retrieves views of the stationary bike, and the answer model answers ``The object you can use to exercise in the TV room is the exercise bike.'' This answer matches the ground truth ``The Stationary Bike'' (score 5; Multi-Frame VLMs: 4; RAG with BLIP: 5).

\noindent\textbf{Limitation in route following.}
\Cref{fig:route} shows a failure case that requires integrating evidence across multiple viewpoints. For the question ``If you were to go down the stairs, then straight into the tv room, turn left in the door and go straight until you enter the last room, which room would it be?'', the ground truth is ``The restroom'', but our method answers ``The room would be the storage room.'' (score 1; Multi-Frame VLMs: 5; RAG with BLIP: 1). The selected views, such as those of the stairs, are individually relevant to the question. However, our method selects each view independently based on relevance and visual redundancy, and it does not model the temporal order of observations or the connectivity between rooms, which are required to follow the described route. This example illustrates the limitation discussed in \cref{sec:limitations} of the main paper.

\section{Fallback Analysis}

\Cref{tab:fallback} reports how often each fallback in Algorithm~\ref{alg:greedy} is triggered, together with the LLM-Match of the questions that trigger it and of the remaining questions. When uniform sampling was replaced with relevance- and diversity-based selection over all candidates for the questions in which no candidate exceeded $\gamma$, the overall LLM-Match decreased from 57.5\% to 56.1\%.

\begin{table}[t]
  \centering
  \caption{Frequency and LLM-Match of each fallback mechanism. $\mathcal{C}_{\mathrm{high}}$ and $\mathcal{C}_{\mathrm{low}}$ denote the candidates whose ITM scores are at least and below $\gamma$, respectively.}
  \label{tab:fallback}
  \begin{tabular}{@{}lccc@{}} \toprule
    Fallback & Questions & Triggered & Not triggered \\ \midrule
    Uniform sampling ($\lvert\mathcal{C}_{\mathrm{high}}\rvert = 0$) & 48 (8.6\%) & 53.1 & 57.9 \\
    Similarity-constraint relaxation & 28 (5.0\%) & 39.3 & 58.5 \\
    Filling from $\mathcal{C}_{\mathrm{low}}$ & 22 (3.9\%) & 23.9 & 58.9 \\ \bottomrule
  \end{tabular}
\end{table}

\section{Relevance Model Comparison}

We compare relevance models under the same threshold-free Top-$K$ protocol, using the same cubemap candidates and answer model (\cref{tab:relevance}). For SigLIP~2~\cite{tschannen2025siglip2}, we use the So400m model without fine-tuning and rank candidates by the cosine similarity between image and question embeddings. This comparison replaces only the relevance model and does not apply Diversity-Aware Greedy Selection.

\begin{table}[t]
  \centering
  \caption{Top-$K$ selection with different relevance models.}
  \label{tab:relevance}
  \begin{tabular}{@{}lc@{}} \toprule
    Relevance model & LLM-Match \\ \midrule
    Fine-tuned BLIP-2 (ours) & 55.2 \\
    Pretrained BLIP-2 & 41.7 \\
    Pretrained SigLIP~2 So400m & 27.4 \\ \bottomrule
  \end{tabular}
\end{table}

\section{Observation Units and Runtime Breakdown}

Each omnidirectional observation yields six cubemap candidate views, so the \textit{Original} and \textit{w/o rotation} settings contain on average 1,022.85 and 352.58 candidate views per question (170.48 and 58.76 observations in \cref{tab:avg_obs_frames}). In both settings, the answer model receives $K=5$ selected views, corresponding to 1,620 Qwen2-VL visual tokens (324 per view).

\Cref{tab:breakdown} reports the runtime of each component of the viewpoint selection stage and of answer generation, measured in a separate profiling run on four NVIDIA H100 GPUs with a batch size of 32 per GPU. Table~\ref{tab:time} reports the selection time from a different run with the same batch size; this breakdown additionally measures each component, including image decoding and preprocessing, so the totals differ slightly.

\begin{table}[t]
  \centering
  \caption{Runtime of each stage (seconds per question), measured on four NVIDIA H100 GPUs with a batch size of 32 per GPU.}
  \label{tab:breakdown}
  \begin{tabular}{@{}lc@{}} \toprule
    Component & Time (s) \\ \midrule
    Image decoding & 8.03 \\
    Cubemap generation & 3.95 \\
    BLIP-2 preprocessing & 4.98 \\
    BLIP-2 relevance estimation & 8.20 \\
    Candidate embedding extraction & 0.90 \\
    Diversity-Aware Greedy Selection & 0.001 \\
    Materialization of selected views & 0.60 \\ \midrule
    Total selection stage & 26.67 \\
    Answer generation & 2.72 \\ \midrule
    End-to-end total & 29.38 \\ \bottomrule
  \end{tabular}
\end{table}

\end{document}